\documentclass[10pt,twocolumn,letterpaper]{article}

\usepackage[pagenumbers]{cvpr} 

\usepackage[dvipsnames]{xcolor}

\usepackage{multirow}
\usepackage{placeins}
\usepackage{color, colortbl}
\definecolor{mygray}{gray}{0.926}

\definecolor{cvprblue}{rgb}{0.21,0.49,0.74}
\usepackage[pagebackref,breaklinks,colorlinks,citecolor=cvprblue]{hyperref}

\def\paperID{6829} 
\def\confName{CVPR}
\def\confYear{2024}

\title{Codebook Transfer with Part-of-Speech for Vector-Quantized Image Modeling}

\author{\normalsize Baoquan Zhang$^1$, Huaibin Wang$^1$, Chuyao Luo$^1$, Xutao Li$^1$, Guotao Liang$^1$, Yunming Ye$^1$, Xiaochen Qi$^2$, Yao He$^2$\\
	\normalsize $^1$ Harbin Institute of Technology, Shenzhen; $^2$ SIFAR \\
	{\tt\small baoquanzhang@hit.edu.cn, 22S051022@stu.hit.edu.cn, luochuyao.dalian@gmail.com,}\\
	{\tt\small lianggt@pcl.ac.cn, \{lixutao, yeyunming\}@hit.edu.cn, \{joeqxc1974, heyao18818\}@gmail.com}
}

\begin{document}
\maketitle
\begin{abstract}
Vector-Quantized Image Modeling (VQIM) is a fundamental research problem in image synthesis, which aims to represent an image with a discrete token sequence. Existing studies effectively address this problem by learning a discrete codebook from scratch and in a code-independent manner to quantize continuous representations into discrete tokens. However, learning a codebook from scratch and in a code-independent manner is highly challenging, which may be a key reason causing codebook collapse, i.e., some code vectors can rarely be optimized without regard to the relationship between codes and good codebook priors such that die off finally. In this paper, inspired by pretrained language models, we find that these language models have actually pretrained a superior codebook via a large number of text corpus, but such information is rarely exploited in VQIM. To this end, we propose a novel codebook transfer framework with part-of-speech, called VQCT, which aims to transfer a well-trained codebook from pretrained language models to VQIM for robust codebook learning. Specifically, we first introduce a pretrained codebook from language models and part-of-speech knowledge as priors. Then, we construct a vision-related codebook with these priors for achieving codebook transfer. Finally, a novel codebook transfer network is designed to exploit abundant semantic relationships between codes contained in pretrained codebooks for robust VQIM codebook learning.
Experimental results on four datasets show that our VQCT method achieves superior VQIM performance over previous state-of-the-art methods.  
\end{abstract}    
\section{Introduction}
\label{sec:intro}

With the development of multi-modal learning on representation and generation, unifying all modality with transformer has attracted increasing interest on computer vision and multi-modal domains \cite{girdhar2023imagebind, zhu2023minigpt, zhang2023llama}. It is well-known that transformer \cite{vaswani2017attention} is proposed for modeling discrete token sequence data like text, which is very difficult to directly apply on continuous image data. To address this problem, Vector-Quantized Image Modeling (VQIM) is proposed and has received wide attention recently. VQIM, as a fundamental problem in machine learning, aims to encode an image with a discrete token sequence similar to texts \cite{van2017neural, peng2021generating}.   

\begin{figure}
  \centering
  \begin{subfigure}{0.49\linewidth}
  \includegraphics[width=1.0\columnwidth]{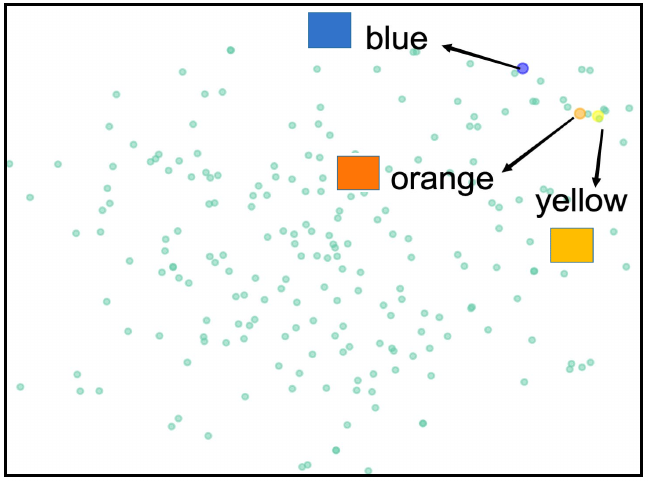}
    \caption{Adjective Codebook Space.}
    \label{fig:short-a}
  \end{subfigure}
  \hfill
  \begin{subfigure}{0.49\linewidth}
    \includegraphics[width=1.0\columnwidth]{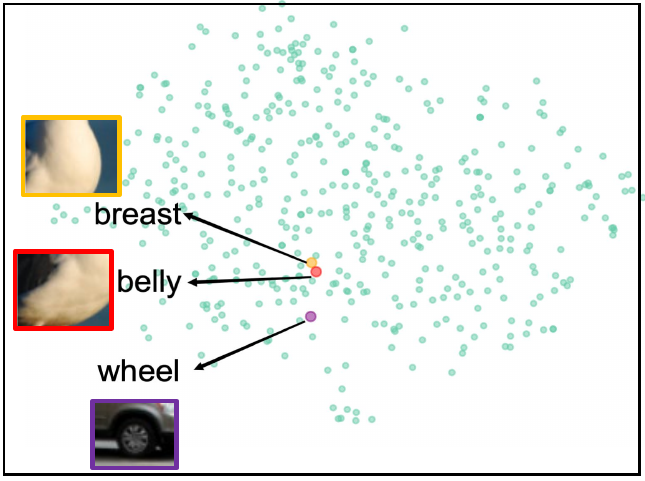}
    \caption{Noun Codebook Space.}
    \label{fig:short-b}
  \end{subfigure}
  \vspace{-8pt}
  \caption{Existing language models actually have provided a superior codebook, which contains abundant semantic relationships between codes and where some vision-related tokens (adjective and noun) can be well transferred to VQIM. For example, the vision-similar ``orange'' and ``yellow'' are indeed closer than dissimilar ``yellow'' and ``blue'' in adjective space (a); In noun space (b), some vision-similar parts like ``breast'' and ``belly'' are also closer than dissimilar ``breast'' and ``wheel''. Resorting to such relationships, VQIM codebook collapse can be alleviated, \emph{e.g.}, although the ”orange” code is not selected to optimize, but its code vector can also be well learned with its close relationship to the ”yellow” code.}
  \vspace{-15pt}
  \label{fig1}
\end{figure}

Existing studies \cite{van2017neural, esser2021taming, zhang2023regularized, zheng2023online, ding2021cogview} effectively address this VQIM problem by learning a discrete codebook from scratch and in a code-independent manner to quantize continuous feature representation into a discrete token sequence. 
For example, in \cite{van2017neural}, Oord et al. propose an encoder-decoder network (called VQ-VAE) to learn and quantize a latent feature space by selecting its nearest neighbor in the codebook as the discrete vector (\emph{i.e.}, token) and training the VQ-VAE model by a simple reconstruction loss. Esser et al. \cite{esser2021taming} further enhance the VQ-VAE by additionally introducing an adversarial loss. Although these methods have shown superior performance on VQIM, they suffer from a codebook collapse issue \cite{roy2018theory}, \emph{i.e.}, only a few code vectors perform optimization during training, whereas a majority of them are never updated (\emph{i.e.}, “die off”). This limits the VQIM performance in these existing methods. To address this codebook collapse issue, recently, some new VQIM techniques are developed from the perspective of codebook update \cite{razavi2019generating, williams2020hierarchical, dhariwal2020jukebox, zheng2023online}, quantization \cite{baevski2019vq, takida2022sq, vuong2023vector}, or regularization \cite{zhang2023regularized} for robust codebook learning. 

In this paper, we also focus on robust codebook learning but present a new perspective (\emph{i.e.}, codebook transfer) for alleviating codebook collapse. Our insight is that neglecting the relationship between code vectors and codebook priors to learn a discrete codebook from scratch is actually very difficult, which may be a key reason causing codebook collapse, \emph{i.e.}, some code vectors can rarely be optimized without regard to the relationship between codes and good codebook priors such that die off finally. Inspired by recent pretrained language models \cite{kenton2019bert, radford2019language, radford2021learning}, we find that some superior codebooks have been well pretrained in some models (\emph{e.g.}, CLIP \cite{radford2021learning} and GloVe \cite{pennington2014glove}) and these code vectors in the codebook are not fully independent, which contains abundant transferable relationships between codes from language to vision \cite{zhang2021prototype, xing2019adaptive}. For example, as shown in Figure~\ref{fig1}, 1) the vision-similar ``orange'' and ``yellow'' are indeed closer than vision-dissimilar ``yellow'' and ``blue'' in adjective codebook space; and 2) two parts (``breast'' and ``belly'') with similar vision  are also closer than ``belly'' and ``wheel'' with dissimilar vision in noun codebook space. Resorting to such transferable relationships, VQIM codebook can be well learned by code cooperative optimization for alleviating the codebook collapse, \emph{e.g.}, although the "orange" code is not selected to optimize, but its code vector can also be well learned by resorting to its relationship with the "yellow" code (see Figure~\ref{fig3} for more details). 

Based on this idea, we propose a novel codebook transfer framework with part-of-speech, which transfers the abundant semantic knowledge of codebook from pretrained language models in order to enhance VQIM codebook learning, called VQCT. Specifically, we first introduce a pretrained language model and part-of-speech knowledge (e.g., WordNet) as priors. Then, we construct a set of vision-related codebook (\emph{i.e.}, adjective and noun codebooks) from the pretrained codebook of language models by filtering out vision-unrelated tokens according to their part-of-speech. Finally, a novel graph convolution-based codebook transfer networks is designed to model the VQIM codebook in a transfer mapping manner from language to vision, which aims to resort to the abundant knowledge from pretrained codebooks to enhance the VQIM codebook learning. The advantage of such design is that 1) the abundant knowledge from well-pretrained codebooks can be fully exploited for providing good codebook priors; and 2) our codebook is generative but not directly learnable such that the semantic relationships between codes can be fully exploited for achieving cooperative optimization between codes. 


Our main contributions can be summarized as follows:
\begin{itemize}
   \item We propose a new perspective, \emph{i.e.}, codebook transfer from language models to VQIM, to alleviate the codebook collapse issue. Its advantage is that the abundant transferable relationships from language codebooks can be fully exploited for enhancing codebook learning.
	
   \item Resorting to part-of-speech knowledge, we construct a set of vision-related codebooks (i.e., adjective and noun codebooks) and design a novel graph convolution-based codebook transfer network. In particular, our codebook is generative rather than directly optimized. Its advantage is that cooperative optimization between codes can be achieved for alleviating codebook collapse issue.
	
   \item We conduct comprehensive experiments on four datasets, which verify the effectiveness of our VQCT method.
\end{itemize}

\vspace{-5pt}
\section{Related Works}
\label{sec:formatting}

\subsection{Vector-Quantized Image Modeling}
Vector-Quantized Image Modeling (VQIM) is a challenging machine learning task, which aims to encode an image with a discrete token sequence like a text token sequence \cite{huang2023not, ning2023all, zheng2022movq, gu2022vector, chang2022maskgit, dong2023peco, lee2022autoregressive}. To address this problem, a large number of VQIM methods have been proposed, which aim to learn a discrete codebook from scratch and in a code-independent manner to quantize continuous representation into a discrete token sequence \cite{van2017neural, esser2021taming, zhang2023regularized, zheng2023online}. Specifiaclly, Oord et al. \cite{van2017neural} first achieve a VQIM model, called VQ-VAE, by following the framework of Variational Auto-Encode (VAE) and replacing its prior dirtribution with a discrete deterministic distribution (i.e., a codebook). Based on the superiority of VQ-VAE, a large number of studies \cite{esser2021taming, zhang2023regularized, zheng2023online, yu2021vector, huang2023towards} further improve its VQIM performance and produce a series of VQ-VAE variants. For example, in \cite{esser2021taming}, Esser et al. further propose a VQ-GAN by introducing an adversarial training loss to enhance the generation quality of VQ-VAE. Yu et al. \cite{yu2021vector} design a vision transformer (ViT) to encode image representations for improving the modeling quality of convolution networks in VQIM. Huang et al. \cite{huang2023towards} propose a dynamic quantization VAE (DQ-VAE) for learning a compact code representation in a variable-length manner. Although these methods have shown superior performance, recent studies show that these methods suffer from a codebook collapse issue \cite{zhang2023regularized}, \emph{i.e.}, only a few codes perform optimization during training, whereas a majority of them are never updated (\emph{i.e.}, “die off”). This limits the VQIM performance of these existing methods.

To address this issue, some new VQIM techniques \cite{zhang2023regularized, takida2022sq, zheng2023online} are proposed from the perspective of codebook update, quantization, or regularization. For example, in \cite{takida2022sq}, a stochastically quantized variational autoencoder (SQ-VAE) is proposed from the perspective of quantization, which alleviates the codebook collapse issue in a self-annealed stochastic quantization manner. Zheng et al. \cite{zheng2023online} develop an online codebook learning strategy from the perspective of codebook update, which aims to restart these unused code vectors into activation status. Zhang et al. \cite{zhang2023regularized} design a regularized vector quantization strategy to mitigate codebook collapse issues, \emph{i.e.}, introducing a prior distribution to regularize the codebook utilization for VQIM. In this paper, we also focus on learning a robust codebook for VQIM. However, different from these existing methods, we present a new codebook transfer perspective to address the codebook collapse issue, which effectively exploits abundant transferable relationships between codes to achieve cooperative optimization between codes. 

\subsection{Pretrained Language Models}
Pretrained Language Models (PLM) \cite{kenton2019bert} have experienced tremendous success with a large number of text corpus, which aims to learn a language understanding and generation model in a self-supervised manner \cite{bao2021beit}. Early models, such as Word2Vector \cite{mikolov2013efficient} and GloVe \cite{pennington2014glove}, laid the foundation for this progress, which focuses on learning a good representation in a self-supervised manner for each word (\emph{i.e.}, token) and then leveraging it to understand/generate text data. With the superiority of transformer, a large number of end-to-end pretained language or vision-language models, e.g., BERT \cite{kenton2019bert}, GPT \cite{radford2019language}, CLIP \cite{radford2021learning}, and ImageBind \cite{girdhar2023imagebind}, are proposed with a large number of text or text-image pair corpora, which demonstrates significant breakthroughs across numerous language or multi-modal tasks. For example, in \cite{radford2021learning}, Radford et al. propose a Contrastive Language-Image Pre-Training (CLIP) to learn a robust representation for each text and image. In \cite{kenton2019bert}, Devlin et al. design a new language representation model (i.e., Bidirectional Encoder Representations from Transformers, called BERT), and then learn it in a masked language model manner. 

In this paper, we find that the pretrained word embeddings (\emph{i.e.}, tokens) from PLM have actually provided a good codebook prior 
and propose a codebook transfer framework to enhance VQIM codebook learning. We note that a concurrent work with our VQCT is SPAE \cite{yu2023spae}. However, different from SPAE that directly regarding the pretrained codebook from PLM as VQIM codebook, we target at transfering the pretrained codebook from PLM to VQIM, and carefully design a novel codebook transfer network with part-of-speech. Its advantage is abundant semantic relationships from pretrained codebooks can be fully exploited for cooperative optimization between codes. 



\section{Methodology}
\subsection{Preliminaries: VQ-VAE}
VQ-VAE \cite{van2017neural}, as a pioneering work on VQIM, aims to learn an encoder-decoder network and a discrete codebook to quantize an image into a discrete token sequence. Formally, let $f_{\theta_{e}}(\cdot)$ with parameter $\theta_{e}$, $f_{\theta_d}(\cdot)$ with parameter $\theta_{d}$, and $\mathcal{C}=\{(k, e_k \in \mathrm{R}^{n_c})\}_{k=1}^{K}$ with parameter $\{e_k \in \mathrm{R}^{n_c}\}_{k=1}^{K}$ denote the encoder, decoder, and discrete codebook, respectively. Among them, $K$ is the size of the codebook, $e_k$ denotes the $k-$th code vector in the codebook, and $n_c$ is the vector dimension. Given an image $x \in \mathrm{R}^{W \times H \times C}$, we first leverage the encoder $f_{\theta_{e}}(\cdot)$ to encode it into a set of continuous feature vectors $Z \in \mathrm{R}^{w \times h \times n_c}$, \emph{i.e.}, $Z = f_{\theta_{e}}(x)$. Then, a quantization operation $q(\cdot)$ is employed to quantize each continuous feature vector $z \in Z$ into a discrete code sequence, \emph{i.e.}, selecting its nearest neighbor in the codebook $\mathcal{C}$ as its discrete code sequence $D^q$. That is, 
\begin{equation}
    \begin{aligned}
    D^q = q(Z, \mathcal{C}) = \arg\min_{k \in [0, K-1]} \|z - e_k\|_2^2.
    \end{aligned}
    \label{eq_1}
\end{equation}
As a result, a quantized feature $Z^q$ can be obtained, \emph{i.e.}, $Z^q = e_{D^q}$. After that, the quantized feature vector $Z^q$ is fed into the decoder $f_{\theta_d}(\cdot)$ to reconstruct the origin image, \emph{i.e.}, $\hat{x}=f_{\theta_d}(Z^q)$. Finally,  the encoder $f_{\theta_{e}}(\cdot)$, decoder $f_{\theta_d}(\cdot)$, and codebook $\mathcal{C}=\{(k, e_k \in \mathrm{R}^{n_c})\}_{k=1}^{K}$ are jointly learned by minimizing the following loss objective:
\begin{equation}
    \begin{aligned}
    L = & \|x-\hat{x}\|_2^2 + \|sg[f_{\theta_{e}}(x)]-Z^q\|_2^2 \\ & + \|f_{\theta_{e}}(x)-\beta sg[Z^q]\|_2^2,
    \end{aligned}
    \label{eq_2}
\end{equation}
where $sg[\cdot]$ is a stop-gradient operator and $\beta$ is a hyperparameter. The first term is reconstruction loss $L_{rec}$, and others are codebook loss $L_{cod}$ which trains the codebook to represent continuous features. With the superiority of VQ-VAE, recent studies propose some varieties, e.g., VQ-GAN that introduce an adversarial loss $L_{adv}$ to improve image generation quality of VQ-VAE.

Although these methods have shown superior performance, they perform optimization only for the active codes (\emph{i.e.}, these codes $D^q$ selected by the quantization operation $q(\cdot)$) and others remain unchanged. This results in that some code vectors may be never optimized such that die off finally, which is called a \emph{codebook collapse issue}. 

\begin{figure*}
  \centering
  \includegraphics[width=1.0\textwidth]{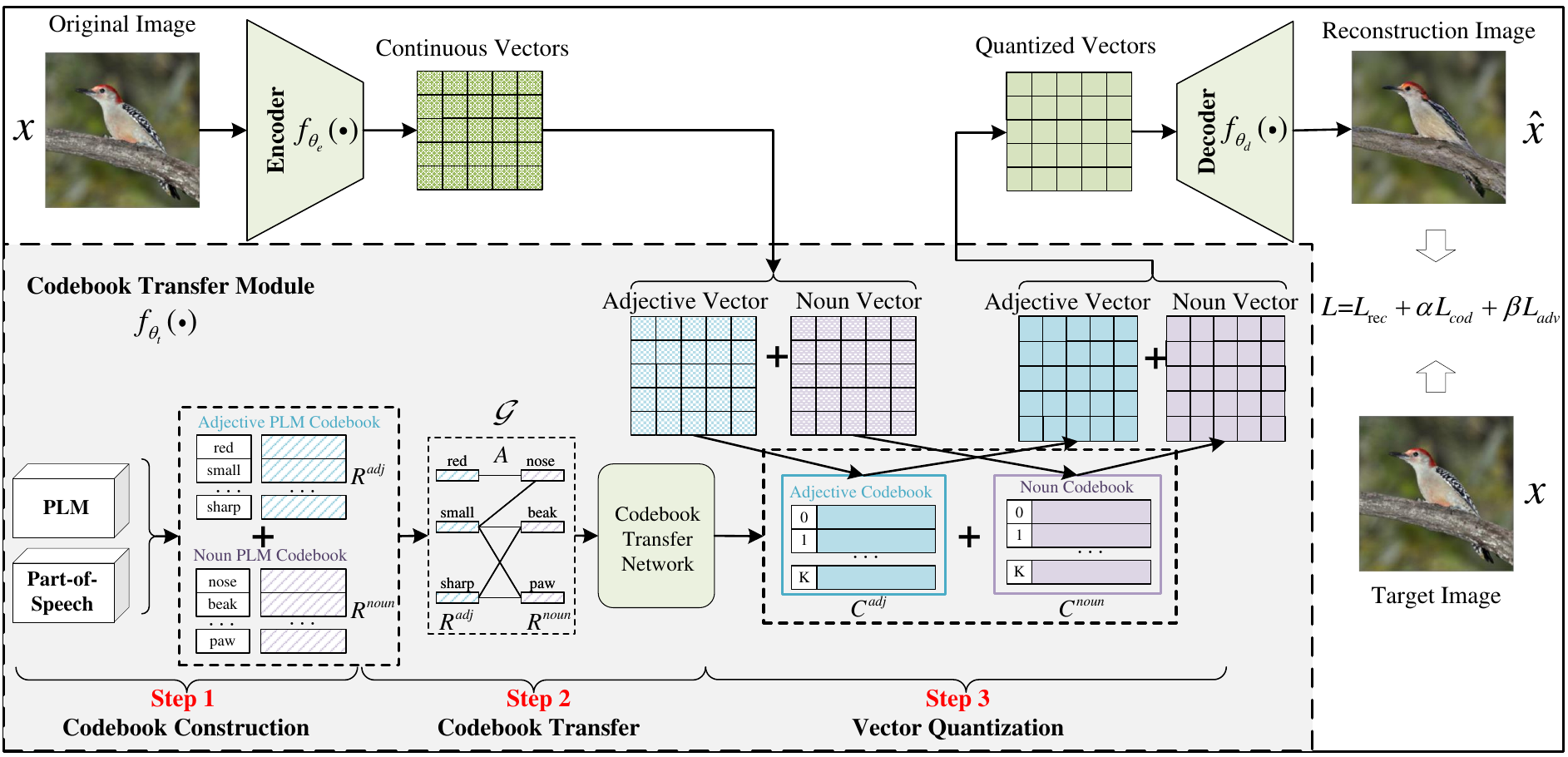}
  \caption{Illustration of our codebook transfer framework with part-of-speech, \emph{i.e.}, VQCT, which consists of an encoder, a codebook transfer module, and a decoder. Here, the encoder aims to represent an image as a set of spatial continuous vectors. Then, the codebook transfer module is employed to generate a codebook in a transfer manner from pretrained language models (PLM) to VQIM and quantize the continuous vector into a set of quantized vectors. Finally, the decoder is used to reconstruct original images with the quantized vectors. 
}
  \label{fig2}
\end{figure*}

\subsection{Overall Framework of VQCT}
Recently, some studies also attempt to address the codebook collapse issue, but most existing methods focus on learning a codebook from scratch and in a code-independent manner. However, in this paper, our insight is neglecting the relationship between code vectors and codebook priors to learn a discrete codebook from scratch is actually very difficult, which may be a key reason causing codebook collapse. Based on this, we propose a novel graph convolution-based code transfer framework with part-of-speech for VQIM, called VQCT. The main idea is that instead of directly learning a codebook from scratch, we introduce a well-pretrained codebook from language models, and part-of-speech knowledge as priors, and then resort to the abundant semantics and relationships contained in these priors to enhance the VQIM codebook learning.

As shown in Figure~\ref{fig2}, our VQCT framework consists of three modules, \emph{i.e.}, an encoder $f_{\theta_e}(\cdot)$ with parameter $\theta_e$, a codebook transfer module $f_{\theta_t}(\cdot)$ with parameter $\theta_t$, and a decoder $f_{\theta_d}(\cdot)$ with parameter $\theta_d$. Given an image $x$, the encoder $f_{\theta_e}(\cdot)$ is employed to encode continuous spatial vectors. Then, instead of directly learning a codebook from scratch, based on the pretrained codebook and part-of-speech priors, the codebook transfer module $f_{\theta_t}(\cdot)$ accounts for predicting a codebook in a transfer manner from language models to VQIM, which is then leveraged to quantize the continuous spatial representations into a discrete vectors. Finally, the decoder $f_{\theta_d}(\cdot)$ is used to reconstruct the original image $\hat{x}$. The workhorse of our VQCT is the codebook transfer network $f_{\theta_t}(\cdot)$. Next, we elaborate on them.

\subsection{Codebook Transfer Module}
Instead of directly learning a codebook from scratch, the codebook transfer module $f_{\theta_t}(\cdot)$ aims to generate a VQIM codebook in a transfer manner from pretrained language models. Its advantage is that abundant semantic relationships between codes can be fully exploited for cooperative learning between codes. Next, we introduce how to generate the VQIM codebook and quantize continuous vectors into quantized vectors, including the following three steps:

\textbf{Step 1: \emph{Codebook Construction.}} In fact, some pretrained language models (e.g., CLIP \cite{radford2021learning}, GloVe \cite{pennington2014glove}, or BERT \cite{kenton2019bert}) have provided superior codebooks (\emph{i.e.}, the embeddings of word tokens), but such information rarely be exploited. Our main idea is transferring these pretrained codebook to VQIM for enhancing VQIM codebook learning. However, transferring all word tokens to VQIM is impractical since the number of word tokens is very large, which imposes a considerable computational burden. Intuitively, a large number of words are all vision-unrelated (e.g., pron., adv., art., prep., or conj.), only few words are vision-related (e.g., adjective and noun). To this end, we first introduce a  well-trained codebook from pretrained language models (PLM) and part-of-speech knowledge (e.g., WordNet) as priors. Then, based on the part-of-speech knowledge, we filter out all vision-unrelated words and only the adjective and noun are retained to construct the vision-related codebook, called ``Adj. PLM Codebook'' and ``Noun. PLM Codebook'', respectively. Let $R^{adj}=\{r^{adj}_i\}_{i=0}^{K_{adj}-1}$ and $R^{noun}=\{r^{noun}_i\}_{i=0}^{K_{noun}-1}$ denote the set of `Adj PLM Codebook'' and `Noun PLM Codebook'', respectively, where $K_{adj}$ and $K_{noun}$ is the code number.


\begin{figure*}
  \centering
  \begin{subfigure}{0.48\linewidth}
  \includegraphics[width=1.0\columnwidth]{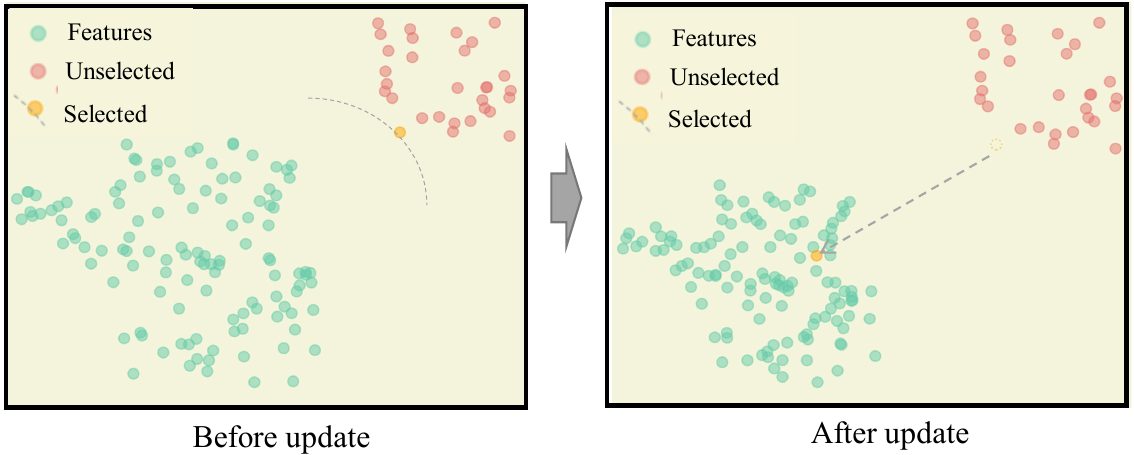}
    \caption{Distribution Change of VQ-VAE}
    \label{fig:short-a}
  \end{subfigure}
  \quad
  \begin{subfigure}{0.48\linewidth}
\includegraphics[width=1.0\columnwidth]{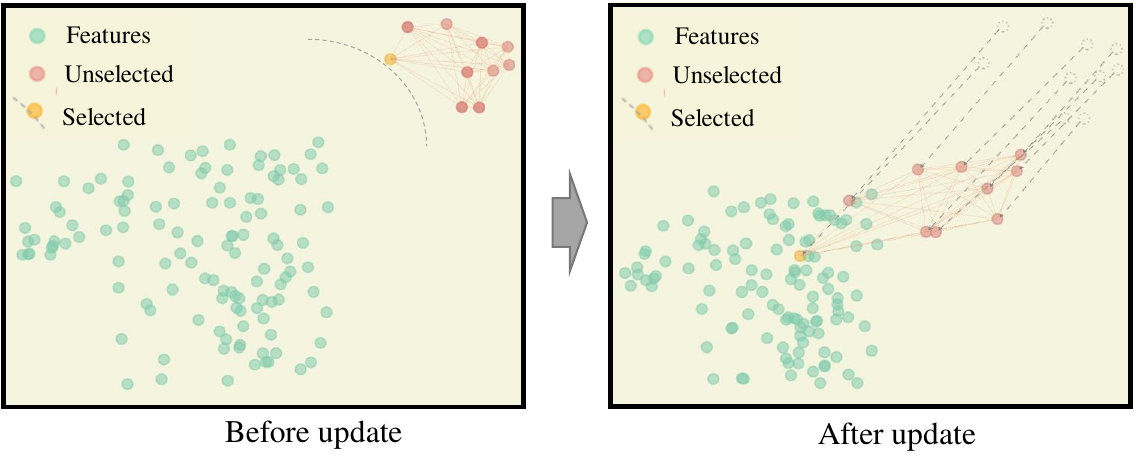}
    \caption{Distribution Change of Our VQCT}
    \label{fig:short-b}
  \end{subfigure}
  \caption{Illustration of codebook optimization. Here, we take a two-dimensional toy setting as an example to show distribution change of codebook when performing optimization. Different from VQ-VAE (a) that only the active “lucky” seeds (in Peach) are optimized but the other “dead” vectors (in Red) are not optimized and remain fixed, our VQCT update all code vectors in the codebook, although only an active code vector is selected in the codebook to perform optimization, with the abundant semantic relationships from pretrained codebook.}
  \label{fig3}
\end{figure*}

\textbf{Step 2: \emph{Codebook Transfer.}} The above ``Adj. PLM codebook'' $R^{adj}$ and ``Noun. PLM codebook'' $R^{noun}$ contains abundant semantics and relationships between word tokens. A simple approach is concatenating these two codebooks ($R^{adj}$ and $R^{noun}$) as a whole codebook, and then directly regarding it as the VQIM codebook or employing a simple multilayer perceptron (MLP) to generate the VQIM codebook. However, the ``Adj. PLM codebbok'' $R^{adj}$ and ``Noun. PLM codebbok'' $R^{adj}$ are not independent, which are jointly used to describe a visual feature, \emph{i.e.}, an adjective is generally used to modify a noun such as ``sharp beak''. 

To achieve this idea, as shown in Figure~\ref{fig2}, we first construct a modifying graph from a large number text corpus to model the modifying relationships between the adjective and noun. Let $\mathcal{G}=(R, A)$ denote the constructed modifying graph where $R=\{R^{adj}, R^{noun}\}$ is node representation from pretrained PLM codebook and $A=\{A_{ij}\}$ denote the modifying relationship matrix between the adjective and noun. In the modifying relationship matrix $A$, $A_{ij} = 1$ if the noun $j$ is modified with the adjective $i$; otherwise $A_{ij} = 0$. Then, we regard the modifying graph $\mathcal{G}=(R, A)$ as inputs and design a graph convolution-based codebook transfer network (GCCTN) $f_{\theta_t}(\cdot)$ to transfer the adjective and noun PLM codebooks to VQIM. That is, 
\begin{equation}
    \begin{aligned}
    (\mathcal{C}_{adj}, \mathcal{C}_{noun}) = f_{\theta_t}(\mathcal{G}),\ \  \mathcal{G}=(R, A),
    \end{aligned}
    \label{eq_2}
\end{equation}
where the GCCTN network consists of three graph convolution layers, which is followed by ReLU activation in each layer, respectively. As a result, a set of VQIM codebooks (\emph{i.e.}, an adjective codebook and a noun codebook) is obtained, which is denoted by $\mathcal{C}_{adj}=\{(k, e_k \in \mathrm{R}^{n_c})\}_{k=1}^{K_{adj}}$ and $\mathcal{C}_{noun}=\{(k, e_k \in \mathrm{R}^{n_c})\}_{k=1}^{K_{noun}}$, respectively. 

\textbf{Step 3: \emph{Vector Quantization.}} Given an image $x$, we first leverage the encoder $f_{\theta_e}(\cdot)$ to obtain its continuous vectors, which is divided into two parts along its channels, \emph{i.e.}, an adjective vector and a noun vector. Then, based on the adjective codebook $\mathcal{C}_{adj}$ and noun codebook $\mathcal{C}_{noun}$, we emply a quantization operation $q(\cdot)$ to quantize each continuous adjective/noun vector into a discrete code sequence, respectively, \emph{i.e.}, selecting its nearest neighbor in the adjective codebook $\mathcal{C}_{adj}$ and noun codebook $\mathcal{C}_{noun}$ as its discrete code sequence (see Eq.~\ref{eq_1}), respectively. After that, a set of quantized adjective vectors and noun vectors can be obtained, which is futher concatenated as quantized vectors. Finally, we feed the quantized vectors into the decoder $f_{\theta_d}(\cdot)$ for reconstructing the origin image $\hat{x}$.

Note that our VQCT only focus on codebook learning, which can be easily integrated into some existing VQIM methods. For example, our VQCT will become 1)  VQCT-VQ-VAE when we adopt Eq.~\ref{eq_2} to train our model; and 2) VQCT-VQ-GAN by adding an adversarial loss on 1). In next sections, we take VQCT-VQ-GAN as the main method to conduct experiments due to its superiority on image synthesis, which is simply called as VQCT.

\subsection{Analysis of Codebook Optimization}
To analyze how our VQCT alleviate codebook collapse issue, we take a two-dimensional codebook setting as a toy example and then visualize the process of codebook optimization in Figure~\ref{fig3}. For clarity, we only show the optimization process of a code vector. From Figure~\ref{fig3}, we can see that only an active (\emph{i.e.}, selected) code vector can be optimized whereas others are never updated (\emph{i.e.}, “die off”) for VQ-VAE, however our VQCT performs cooperative optimization between codes for VQIM, \emph{i.e.}, all codes in our codebook can be performed optimization by resortting to semantic relationship between codes, although only a code vector is selected to learn. This is because 1) our codebook is generative but not directly learnable and 2) our optimization variable is the parameter $\theta_t$ of codebook transfer network instead of the codebook, such that all codes can achieve optimization with their semantic relationships.
\renewcommand\arraystretch{1.25}
\begin{table*}[t]
\footnotesize
\caption{
Results of image reconstruction on ADE20K, CelebA-HQ, CUB-200, and MS-COCO. The best results are highlighted in bold.
}
\renewcommand\tabcolsep{2.25pt}
\centering 
\begin{tabular}{l||cccc||cccc||cccc||cccc} \hline
\multirow{2}{*}{\textbf{Models}} &  \multicolumn{4}{c||}{\textbf{ADE20K \cite{zhou2017scene}}} &  \multicolumn{4}{c||}{\textbf{CelebA-HQ \cite{liu2015deep}}} & \multicolumn{4}{c||}{\textbf{CUB-200 \cite{wah2011caltech}}} &  \multicolumn{4}{c}{\textbf{MS-COCO \cite{lin2014microsoft}}}  \\
\cline{2-17}
& FID$\downarrow$ & PSNR$\uparrow$ & $l1$$\downarrow$ & $l2$$\downarrow$ & FID$\downarrow$ & PSNR$\uparrow$ & $l1$$\downarrow$ & $l2$$\downarrow$ & FID$\downarrow$ & PSNR$\uparrow$ & $l1$$\downarrow$ & $l2$$\downarrow$ & FID$\downarrow$ & PSNR$\uparrow$ & $l1$$\downarrow$ & $l2$$\downarrow$ \\
\hline 
\textbf{VQ-VAE} \cite{van2017neural} & 116.85 & 21.08 & 0.1282 & \textbf{0.0368}  & 36.08 & \textbf{25.29}  & 0.0719 & 0.0139 & 54.92& 24.38 & 0.0849 & 0.0183 & 86.21 & \textbf{23.55} & \textbf{0.0933} & \textbf{0.0226} \\
\textbf{VQ-GAN} \cite{esser2021taming} & 22.04 & 20.42 & 0.1290 & 0.0451  & 5.66 & 24.10 & 0.0798 & 0.0175 &  3.63& 22.19 & 0.1051 & 0.0319 & 14.45 & 20.21 & 0.1311 & 0.0475\\

\textbf{Gumbel-VQ} \cite{baevski2019vq} & 24.12 & 20.04 & 0.1359 & 0.0482  & 6.22 & 23.65  & 0.0837 & 0.0194 & 3.45& 22.11 & 0.1048 & 0.0318 & 15.30 & 20.00 & 0.1354 & 0.0488\\

\textbf{CVQ} \cite{zheng2023online} & 33.63 & 19.91 & 0.1379 & 0.0486  & 5.19 & 23.15  & 0.0917 & 0.0214 & 3.61& 22.29 & 0.1034 & 0.0302 & 9.94 & 20.48 & 0.1253 & 0.0443\\


\rowcolor{mygray} \textbf{VQCT(ours)} & \textbf{20.25} & \textbf{21.30} & \textbf{0.1144} & 0.0374 & \textbf{5.02} & 25.18  & \textbf{0.0699} & \textbf{0.0134} & \textbf{2.13} & \textbf{25.35} & \textbf{0.0704} & \textbf{0.0160} & \textbf{9.82} & 21.46 & 0.1108 & 0.0366 \\
\hline
  
\end{tabular}
\label{tab_1}
\end{table*}

\begin{figure*}[t]
\centering
\includegraphics[width=0.90\linewidth]{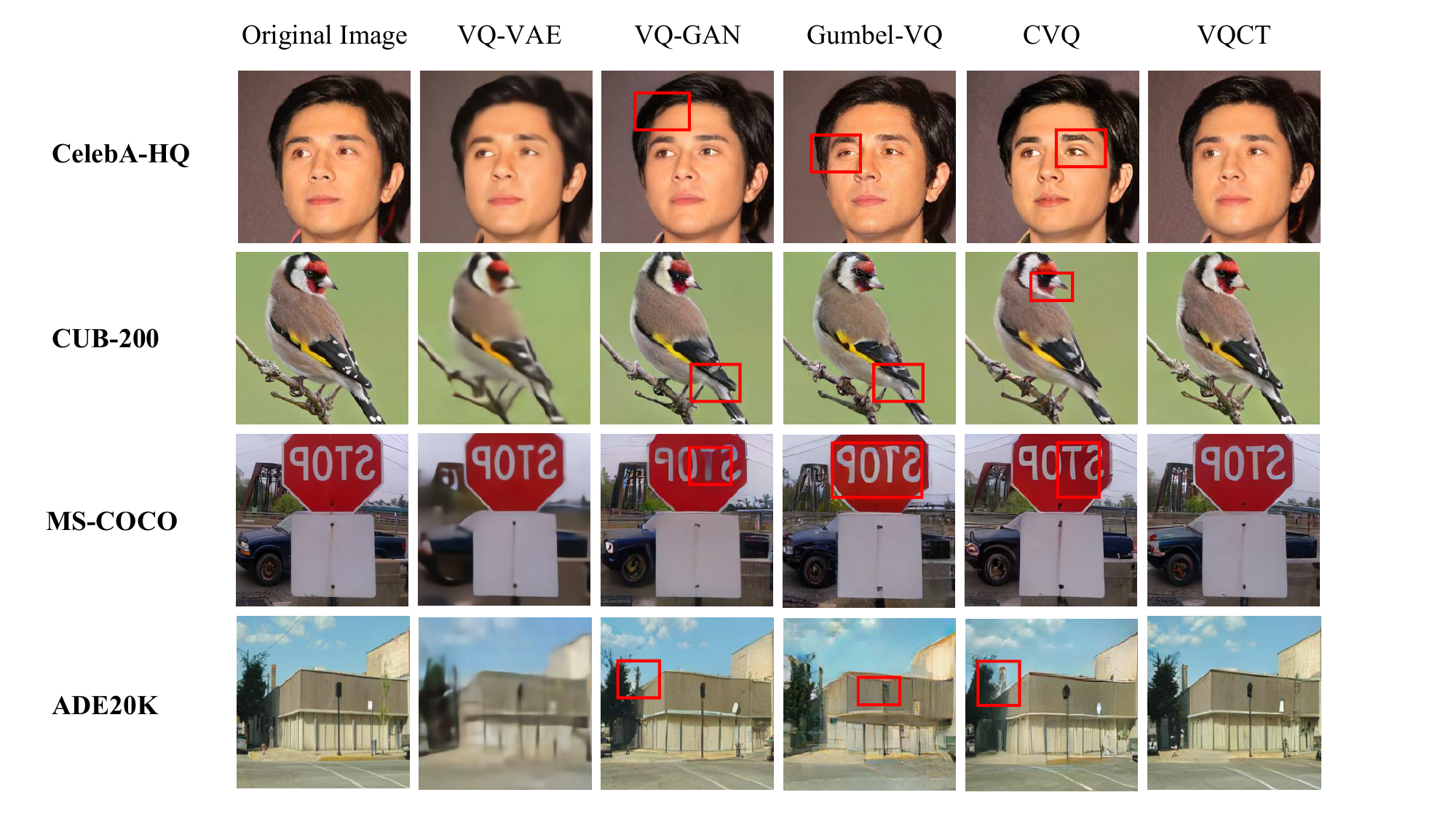}
\caption{
Reconstructed images from different VQIM methods on four datasets. Here, the red-color boxes highlight reconstruction details.
}
\vspace{-15pt}
\label{im_rec}
\end{figure*}

\section{Experiments}
\subsection{Experimental Settings}
\noindent \textbf{Dataset.} We evaluate our method over four public datasets, including ADE20K \cite{zhou2017scene}, CelebA-HQ \cite{liu2015deep}, CUB-200 \cite{wah2011caltech}, and MSCOCO \cite{lin2014microsoft}. Following \cite{zhang2023regularized}, we resize each image as 256 $\times$ 256 resolution for experiment evaluation.

\noindent \textbf{Evaluation Metrics.} We select recent VQ-VAE \cite{van2017neural}, VQGAN \cite{esser2021taming}, Gumbel-VQ \cite{baevski2019vq}, and CVQ \cite{zheng2023online} as our baselines and evaluate these models with image reconstruction performance, which includes four evaluation metrics, \emph{i.e.} Frechet Inception Score (FID) \cite{heusel2017gans} for showing the perceptual similarity of reconstructed images, and $l1$, $l2$, and Peak Signal-to-noise Ratio (PSNR) \cite{fardo2016formal} is employed to measure the pixel-level similarity of image reconstruction. 

\noindent \textbf{Implementation Details.} In our experiments, all parameter settings of our VQCT are same with these existing VQIM methods (e.g., VQVAE, VQGAN, and CVQ) for fair comparison. The key difference lies in that the output of encoder and the input of decoder. We divide the output of encoder into two parts, \emph{i.e.}, an adjective vector and a noun vector, and then concatenate the quantized adjective and noun vectors as the inputs of decoder for reconstruction.

\subsection{Discussion of Results}
\noindent \textbf{Quantitative Evaluation.} In Table~\ref{tab_1}, we select some state-of-the-art VQIM methods (\emph{i.e.}, VQ-VAE, VQ-GAN, Gumbel-VQ, and CVQ) as our baselines and then report the performance of their image reconstruction. It can be found that our VQCT method outperforms these state-of-the-art methods on most evaluation, which suggests that our VQCT method is effective for VQIM. Specifically, compared with VQ-VAE and VQ-GAN, our method better models the codebook by introducing a well-pretrained codebook from language models and word class knowledge as priors. The results show our method is more effective, especially on FID. It's worth noting that our method also beats Gumbel-VQ, and CVQ,
which also focus on alleviate the codebook collapse issue. This shows the effectiveness of our VQCT.

\noindent \textbf{Qualitative Evaluation.} We qualitatively compare reconstruction performance of our VQCT and baseline methods (\emph{i.e.}, VQ-VAE, VQ-GAN, Gumbel-VQ, and CVQ). The results are shown in Figure~\ref{im_rec}. From these results, we can see that our VQ-CT achieves the best reconstruction quality, which can be well aligned with original images in terms of detailed textures. This further verifies the effectiveness.

\subsection{Ablation Study}
\noindent \textbf{Is introducing the pretrained codebook effective for VQIM?} In Table~\ref{tab_2}, we conduct an ablation study to show the effectiveness of introducing the pretrained codebook. Specially, (\romannumeral1) we implement VQ-GAN as our baseline (i.e. without introducing codebook prior into VQIM); (\romannumeral2) we implement our VQCT but randomly initialize our adj and noun codebook and Multilayer Perceptron (MLP) is employed as our code transfer network; and (\romannumeral3) we implement our VQCT, \emph{i.e.}, introducing the pretrained codebook to enhance codebook learning. From the results, we can see that the performance of (\romannumeral3) outperforms (\romannumeral1) and (\romannumeral2) by a large margin, which is reasonable because the semantic relationships between codes from the pretrained codebook can be fully exploited for providing good codebook priors such that more robust codebook can be learned for VQIM. This suggests that our VQCT (\emph{i.e.}, introducing the pretrained codebook from language models) is very effective for VQIM.

\renewcommand\arraystretch{1.25}
\begin{table}[t!]
\small
\caption{
Ablation study of pretrained codebook on CUB-200.
`RI' denote random initialization of adjective and noun codebook.
}
\renewcommand\tabcolsep{8pt}
\centering 
\smallskip\scalebox{0.80}{
\begin{tabular}{l|lcccc} \hline
\multirow{2}{*}{}
& \multirow{2}{*}{\textbf{Setting}}
& 
\multicolumn{4}{c}{\textbf{CUB-200 \cite{wah2011caltech}}} 
\\
\cline{3-6}
& & FID$\downarrow$ & PSNR$\uparrow$ & $l1$$\downarrow$ & $l2$$\downarrow$\\
\hline 
(\romannumeral1) & \textbf{Baseline}(VQ-GAN)  & 3.63 & 22.19 & 0.1051& 0.0249  \\
(\romannumeral2) & \textbf{Our VQCT} (RI) & 3.45 & 23.34 & 0.0908 & 0.0298 \\
(\romannumeral2) & \textbf{Our VQCT} & 2.13 & 25.35 & 0.0704 & 0.0160  \\
  \hline
\end{tabular}}
\label{tab_2}
\end{table}

\renewcommand\arraystretch{1.25}
\begin{table}[t!]
\small
\caption{
Ablation study of our codebook transfer network on CUB-200. Here, `SCB' denote concatenating adj and noun codebooks as a codebook. `MCB' denote regarding adj and noun codebooks as two codebooks, respectively. ``MLP'' denotes implementing our codebook transfer network with MLP. ``GCN'' denotes implementing the our codebook transfer network with GCN. }
\renewcommand\tabcolsep{6pt}
\centering 
\smallskip\scalebox{0.85}{
\begin{tabular}{l|lcccc} \hline
\multirow{2}{*}{}
& \multirow{2}{*}{\textbf{Setting}}
& 
\multicolumn{4}{c}{\textbf{CUB-200 \cite{wah2011caltech}}} 
\\
\cline{3-6}
& & FID$\downarrow$ & PSNR$\uparrow$ & $l1$$\downarrow$ & $l2$$\downarrow$\\
\hline 
(\romannumeral1) & \textbf{Baseline}  & 20.06 & 16.94 & 0.2001& 0.0902  \\
(\romannumeral2) & \textbf{+Finetune} & 3.98 & 22.42 & 0.1036 & 0.0294 \\
(\romannumeral3) & \textbf{+SCB+MLP}  & 5.64 & 21.18& 0.1147& 0.0386  \\
(\romannumeral4) & \textbf{+SCB+GCN} & 3.71 & 22.72 & 0.0973 & 0.0283 \\
(\romannumeral5) & \textbf{+MCB+MLP} & 2.33 & 24.87 & 0.0743& 0.0178  \\
(\romannumeral6) & \textbf{+MCB+GCN (VQCCT)} & 2.13 & 25.35 & 0.0704 & 0.0160  \\
  \hline
\end{tabular}}
\label{tab_3}
\end{table}

\noindent \textbf{Is our codebook transfer network effective?} In Table~\ref{tab_3}, we analyzed the effectiveness of our codebook transfer network. Specifically, (\romannumeral1) we first directly concatenate the pretrained adj and noun embedding as the codebook of VQIM which is freezed, i.e., our baseline (can be regarded as SPAE \cite{yu2023spae}); (\romannumeral2) we further finetune the codebook instead of freezing codebook on (\romannumeral1); (\romannumeral3) we replace finetuning codebook with our codebook transfer on (\romannumeral2) and implement the codebook transfer network with MLP; (\romannumeral4) we implement the codebook transfer network with Graph Convolutional Network (GCN) on (\romannumeral3); (\romannumeral5) we split the adj and noun codebook as two codebooks on (\romannumeral3); (\romannumeral6) we split the adj and noun codebook as two codebooks on (\romannumeral4), which is exactly our VQCT. From the results of (\romannumeral1) $\sim$ (\romannumeral6), we observe that: 1) the performance of (\romannumeral4) $\sim$ (\romannumeral6) exceeds (\romannumeral1) $\sim$ (\romannumeral2) by a large margin, which means that it is helpful to employ a codebook transfer network to transfer pretrained codebook to VQIM for robust codebook learning; 2) the performance of (\romannumeral4)/(\romannumeral6) outperforms (\romannumeral3)/(\romannumeral5), which shows the superiority of using GCN to model the relationship between adjective and noun codebooks. This is because the relationships from adjective and noun part-of-speech can be fully exploited for codebook learning. Finally, comparing the results of (\romannumeral5) $\sim$ (\romannumeral6) with (\romannumeral3) $\sim$ (\romannumeral4), we find that the performance of modeling adj and noun codebook as two codebooks is more superior than a single codebook. This is reasonable because the adj and noun are often used together to describe vision features.

\renewcommand\arraystretch{1.25}
\begin{table}[t!]
\small
\caption{
Ablation study of VQCT generalization on CUB-200. }
\renewcommand\tabcolsep{10pt}
\centering
\smallskip\scalebox{0.8}{
\begin{tabular}{l|lcccc} \hline
\multirow{2}{*}{}
& \multirow{2}{*}{\textbf{Setting}}
& 
\multicolumn{4}{c}{\textbf{CUB-200 \cite{wah2011caltech}}} 
\\
\cline{3-6}
& & FID$\downarrow$ & PSNR$\uparrow$ & $l1$$\downarrow$ & $l2$$\downarrow$\\
\hline 
\multirow{2}{*}{\textbf{(\romannumeral1)}} & \textbf{VQ-VAE}  & 54.92 & 24.38 & 0.0849& 0.0183  \\
& \textbf{VQ-VAE+VQCT}  & 24.39 & 25.77 & 0.0720& 0.0128  \\
\hline
\multirow{2}{*}{\textbf{(\romannumeral2)}} & \textbf{VQGAN} & 3.63 & 22.19 & 0.1051 & 0.0319 \\
 & \textbf{VQGAN+VQCT} & 2.13 & 25.35 & 0.0704 & 0.0160  \\
  \hline
\end{tabular}}
\label{tab_4}
\end{table}

\renewcommand\arraystretch{1.25}
\begin{table}[t!]
\small
\caption{
Ablation study of language models on CUB-200. }
\renewcommand\tabcolsep{10pt}
\centering 
\smallskip\scalebox{0.85}{
\begin{tabular}{l|lcccc} \hline
\multirow{2}{*}{}
& \multirow{2}{*}{\textbf{Setting}}
& 
\multicolumn{4}{c}{\textbf{CUB-200 \cite{wah2011caltech}}} 
\\
\cline{3-6}
& & FID$\downarrow$ & PSNR$\uparrow$ & $l1$$\downarrow$ & $l2$$\downarrow$\\
\hline 
(\romannumeral1) & \textbf{Baseline}  & 3.63 & 22.19 & 0.1051& 0.0319  \\
(\romannumeral2) & \textbf{+GloVe}  & 3.43 & 22.99 & 0.0931& 0.0271  \\
(\romannumeral3) & \textbf{+CLIP} & 2.13 & 25.35 & 0.0704 & 0.0160 \\
  \hline
\end{tabular}}
\label{tab_5}
\end{table}

\begin{figure}
  \centering
  \begin{subfigure}{0.32\linewidth}
  \includegraphics[width=1.0\columnwidth]{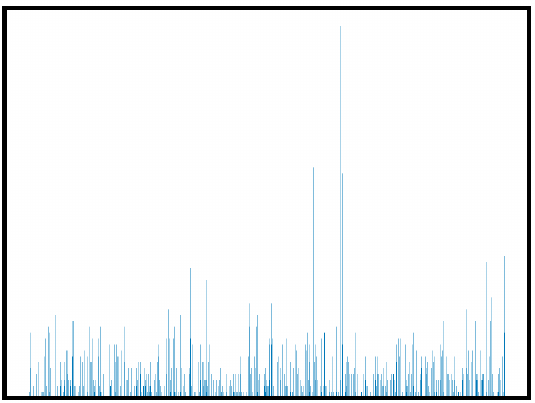}
    \caption{VQGAN}
    \label{fig:short-a}
  \end{subfigure}
  \begin{subfigure}{0.32\linewidth}
  \includegraphics[width=0.99\columnwidth]{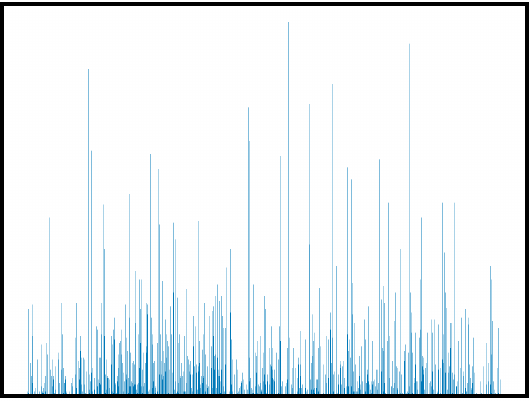}
    \caption{VQCT (Adj)}
    \label{fig:short-a}
  \end{subfigure}
  \begin{subfigure}{0.32\linewidth}
  \includegraphics[width=1.0\columnwidth]{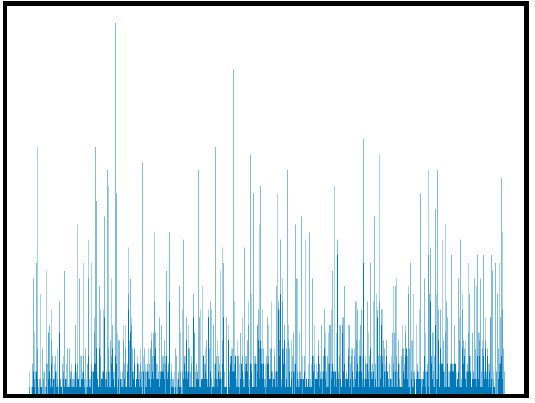}
    \caption{VQCT (Noun)}
    \label{fig:short-a}
  \end{subfigure}
  \caption{Visualization of codebook utilization on CUB-200.}
  \label{fig5}
\end{figure}

\begin{figure}
  \centering
  \begin{subfigure}{0.48\linewidth}
  \includegraphics[width=1.0\columnwidth]{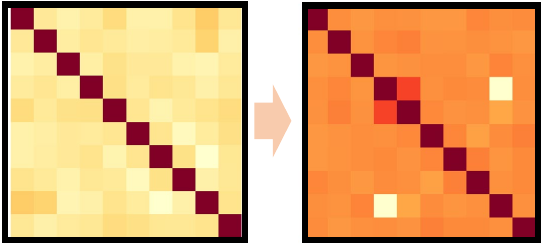}
    \caption{Similarity Change of VQ-GAN}
    \label{fig:short-a}
  \end{subfigure}
  \begin{subfigure}{0.470\linewidth}
  \includegraphics[width=1.0\columnwidth]{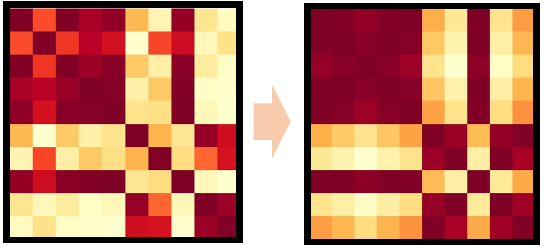}
    \caption{Similarity Change of VQCT}
    \label{fig:short-a}
  \end{subfigure}
  \caption{Visualization of codebook similarity on CUB-200. }
  \label{fig7}
\end{figure}

\begin{figure}
  \centering
  \begin{subfigure}{0.41\linewidth}
  \includegraphics[width=1.0\columnwidth]{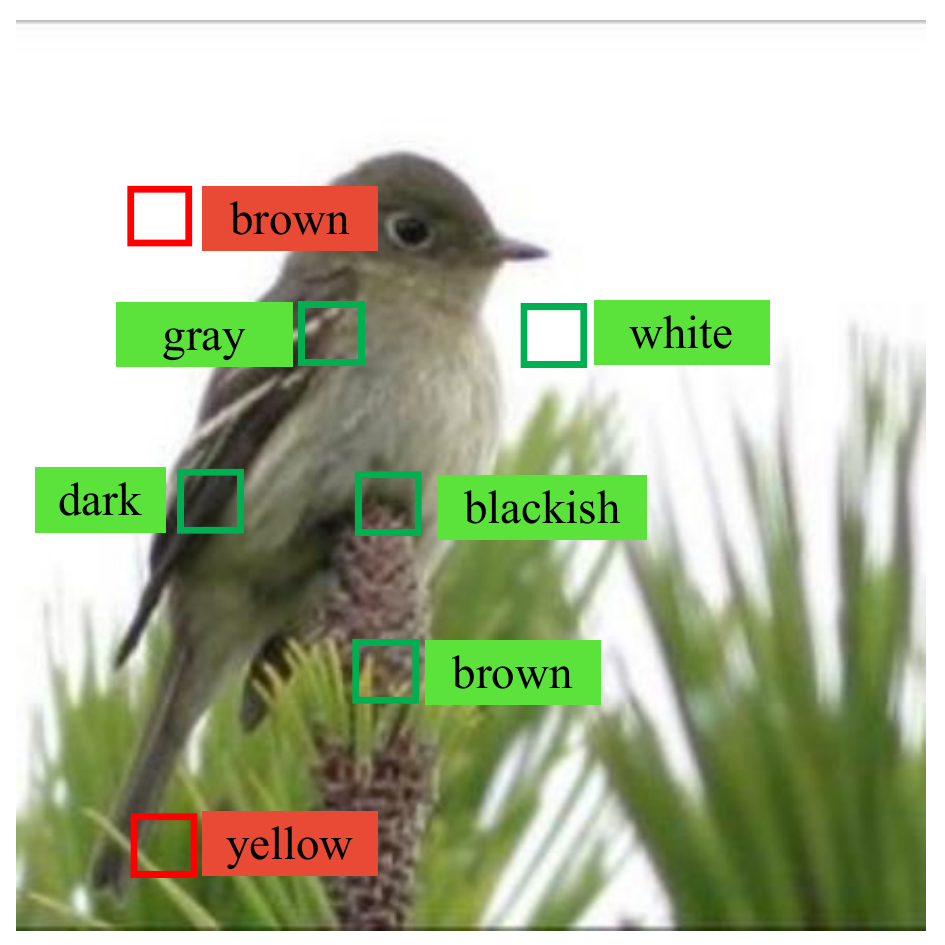}
    \caption{Adjective Space}
    \label{fig:short-a}
  \end{subfigure}
  \quad \quad
  \begin{subfigure}{0.41\linewidth}
  \includegraphics[width=1.0\columnwidth]{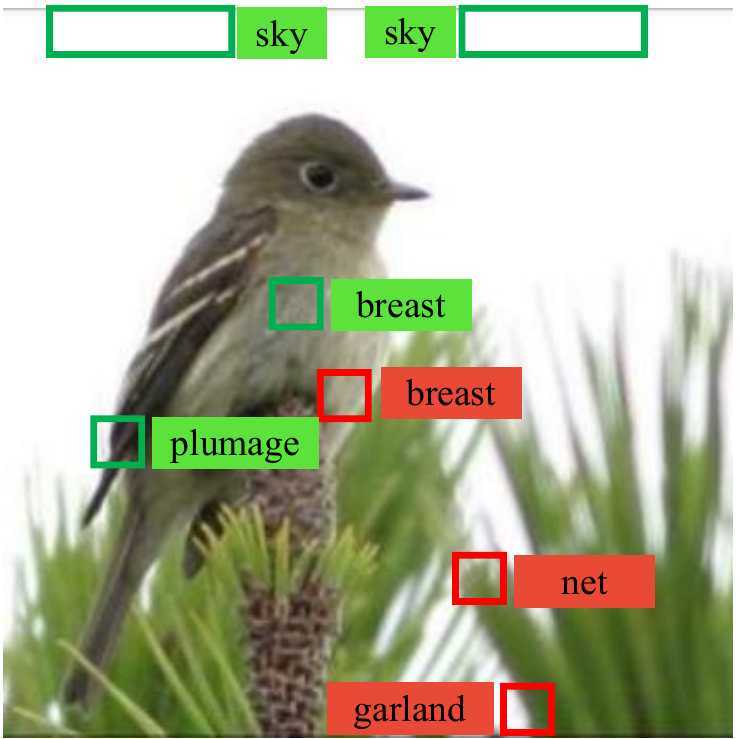}
    \caption{Noun Space}
    \label{fig:short-a}
  \end{subfigure}
  \caption{Visualization of vision-language alignment on CUB-200. Wrong/right matching is marked in red/green, respectively. }
  \label{fig8}
\end{figure}

\noindent \textbf{Is our VQCT general on other VQIM methods?} In fact, our VQCT can be regarded as introducing the codebook transfer strategy into VQ-GAN, which is very general on other VQIM methods. To illustrate this point, we conduct an extension experiment on VQ-VAE and VQ-GAN. Specifically, we first report the performance of image reconstruction with VQ-VAE and VQ-GAN, respectively, and then we add our codebook transfer strategy to enhance the robust codebook learning for VQIM. The experimental results are shown in Table~\ref{tab_4}.  From these results, the VQIM performance of VQ-VAE and VQ-GAN achieve significant performance improvement after applying our codebook transfer strategy to the codebook of VQ-VAE and VQ-GAN, which verifies the universality of our VQCT.

\noindent \textbf{How does pretrained codebook from different language models impact the performance of our VQCT?} In Table~\ref{tab_5}, we analyze the impact of different language models including Glove \cite{pennington2014glove} and CLIP \cite{radford2021learning} on VCQT. Specifically, (\romannumeral1) we first select VQ-GAN as our baseline; (\romannumeral2) 
we introduce a pretrained codebook from the GloVe-based language model on (\romannumeral1) ; (\romannumeral3) we introduce a pretrained codebook from the CLIP-based language model on (\romannumeral1). From these results, we can see that 1) the image reconstruction performance of (\romannumeral2) $\sim$ (\romannumeral3) exceed (\romannumeral1) by a large margin. This is reasonable because we introduce a prerained codebook from language model to enhance codebook learning on (\romannumeral2) $\sim$ (\romannumeral3); and 2) the setting of (\romannumeral3)  (introducing  pretrained codebook from CLIP) performs more superior than other settings. Such advantage may be from the pretraining of multimodal alignment. Hence, CLIP is a default setting in our approach.

\noindent \textbf{Can our VQCT alleviate the codebook collapse issue?} To answer this question, we select VQ-GAN as baseline and then visualize the codebook utilization of the baseline and our VQCT. The visualization result is shown in Figure~\ref{fig5}. From Figure~\ref{fig5}, we can see that 1) in VQ-GAN, the codebook utilization is very low (around 27.90\% codes are used and 72.10\% codes dies off); while 2) the codebook utilization achieves significant improvement, around 42.20\% adj and noun codes is used after applying our VQCT. 

\noindent \textbf{Can our VQCT achieve code cooperative optimization?} In Figure~\ref{fig7}, we select VQ-GAN as baseline and then visualize the cosine similarity between codes. For clarity, we randomly select 10 codes to conduct experiments. We can see that compared with VQ-GAN, the similarity between codes of our VQCT can indeed be well maintained during training, which means that our VQCT can achieve the cooperative optimization between codes.

\noindent \textbf{Can our VQCT align vision and language?} In Figure~\ref{fig8}, we visualize the codes of adjective and noun and mark some right or wrong matching cases. From results, we find that our VQCT can align vision and language in some cases, although by only using unsupervised learning manner, which verifies the effectiveness of our codebook transfer. 

\renewcommand\arraystretch{1.25}
\begin{table}[t]
\footnotesize
\caption{
Results (FID$\downarrow$) of image synthesis on CelebA-HQ \cite{liu2015deep} datasets. The best results are highlighted in bold. 
}
\renewcommand\tabcolsep{2pt}
\centering 
\smallskip\scalebox{0.82}{
\begin{tabular}{cccccc} \hline
\multirow{1}{*}{\textbf{Models}} &  \multicolumn{1}{c}{\textbf{VQ-VAE}} \cite{van2017neural} &  \textbf{VQ-GAN} \cite{esser2021taming} & \textbf{Gumbel-VQ} \cite{baevski2019vq} & \textbf{Reg-VQ} \cite{baevski2019vq} & \textbf{VQCT(ours)} \\
\cline{2-3}
\hline 
FID & 39.57 & 17.42 & 16.78 & 15.34 & \textbf{14.47}\\
\hline
\end{tabular}}
\label{tab_6}
\end{table}

\begin{figure}
\centering
\includegraphics[width=1.0\linewidth]{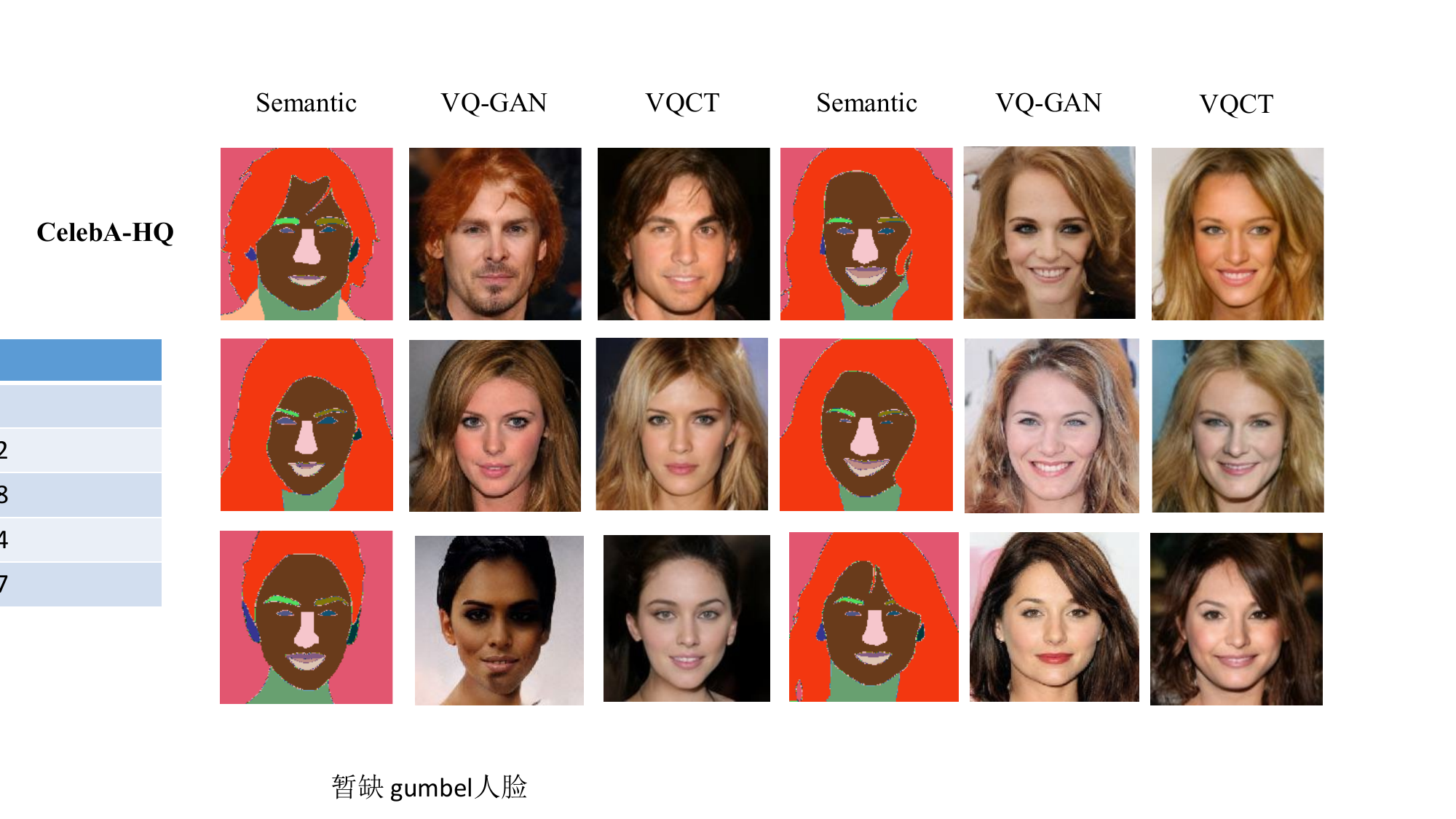}
\caption{Semantic image synthesis on CelebA-HQ.}
\label{fig9}
\end{figure}

\subsection{Application: Image Synthesis}
Following \cite{esser2021taming}, we conduct a simple experiment to verify the effectiveness of our VQCT on downstream tasks, \emph{i.e.}, semantic image synthesis, where LDM \cite{esser2021taming} is employed to achieve the prediction of the adjective and noun codes. 

\noindent \textbf{Quantitative Evaluation.} In Table~\ref{tab_6}, we select latest VQ-VQE, VQ-GAN, Gumbel-VQ, and Reg-VQ as baselines and report the FID performance. From results, we can see that our VQCT achieve superior performance over these baseline methods. This further verifies the effectiveness of our VQCT on downstream semantic image synthesis.

\noindent \textbf{Qualitative Evaluation.} In Figure~\ref{fig9}, we show some generation examples of VQ-GAN and our VQCT on image synthesis and completion.  From results, we can see that our VQCT indeed can achieve high quality image synthesis.

\section{Conclusions}
This paper proposes a novel codebook transfer framework for vector-quantized image modeling. In particular, we introduce a pretrained codebook and part-of-speech knowledge as priors and then design a graph convolution codebook transfer network to generate codebook. Its advantage is rich semantic from pretrained codebook can be fully exploited for codebook learning. Results show that our VQCT achieves superior performance over previous methods. 

{
    \small
    \bibliographystyle{ieeenat_fullname}
    \bibliography{main}
}


\end{document}